\documentclass[letterpaper]{article} 
\usepackage[]{aaai25}  
\usepackage{times}  
\usepackage{helvet}  
\usepackage{courier}  
\usepackage[hyphens]{url}  
\usepackage{graphicx} 
\usepackage{natbib}  

\usepackage{caption} 
\usepackage{algorithm}
\usepackage{algorithmic}
\usepackage{booktabs}
\usepackage{amsfonts}       
\usepackage{nicefrac}       
\usepackage{microtype}      
\usepackage{xcolor}         
\usepackage{amsmath}
\usepackage{amssymb}
\usepackage{subfigure}
\usepackage{multirow}
\DeclareMathOperator*{\myhardamard}{\raisebox{-3.5pt}{\scalebox{1.47}{$\Lambda$}}}
\usepackage{newfloat}
\usepackage{listings}
\DeclareCaptionStyle{ruled}{labelfont=normalfont,labelsep=colon,strut=off} 
\floatstyle{ruled}
\newfloat{listing}{tb}{lst}{}
\floatname{listing}{Listing}
\title{A Wander Through the Multimodal Landscape:\\
Efficient Transfer Learning via Low-rank Sequence Multimodal Adapter}
\author{
    Zirun Guo,
    Xize Cheng,
    Yangyang Wu,
    Tao Jin\thanks{Corresponding author}
}
\affiliations{
    Zhejiang University\\

    \texttt{zrguo.cs@gmail.com}
}

\usepackage{bibentry}

\begin{document}

\maketitle

\begin{abstract}
    Efficient transfer learning methods such as adapter-based methods have shown great success in unimodal models and vision-language models. However, existing methods have two main challenges in fine-tuning multimodal models. Firstly, they are designed for vision-language tasks and fail to extend to situations where there are more than two modalities. Secondly, they exhibit limited exploitation of interactions between modalities and lack efficiency. To address these issues, in this paper, we propose the lo\textbf{W}-r\textbf{a}nk seque\textbf{n}ce multimo\textbf{d}al adapt\textbf{er} (\textbf{Wander}). We first use the outer product to fuse the information from  different modalities in an element-wise way effectively. For efficiency, we use CP decomposition to factorize tensors into rank-one components and achieve substantial parameter reduction. Furthermore,  we implement a token-level low-rank decomposition to extract more fine-grained features and sequence relationships between modalities. With these designs, Wander enables token-level interactions between sequences of different modalities in a parameter-efficient way. We conduct extensive experiments on datasets with different numbers of modalities, where Wander outperforms state-of-the-art efficient transfer learning methods consistently. The results fully demonstrate the effectiveness, efficiency and universality of Wander.
\end{abstract}

\section{Introduction}
In recent years, multimodal models, such as BLIP-2~\citep{li2023blip}, LLaVA~\citep{liu2024visual}, have experienced rapid development and have shown excellent performance in various downstream tasks. However, the increasing number of parameters in these multimodal models, coupled with the diversification of downstream tasks, has resulted in a significant consumption of computational resources and time for fine-tuning these models. Consequently, efficient transfer learning strategies~\citep{hu2021lora, houlsby2019parameter, vu2021spot, liu2023gpt, dettmers2024qlora} have become a focal point of current research.

For multimodal models, there are two kinds of popular efficient transfer learning strategies, including adapter-based methods~\citep{zhang2021tip, sung2022vl, gao2024clip, lu2024uniadapter} and prompt learning methods~\citep{zang2022unified, xing2023dual, khattak2023maple, guo2024multimodal, yan2024low}. Adapter-based methods add extra modules and only train these added parameters while freezing the pre-trained model. For example, \citet{lu2024uniadapter} propose a unified and knowledge-sharing design to interact between the vision and language modalities. Prompt learning methods insert trainable prompts to the input or attention matrices. For instance, \citet{khattak2023maple} propose a coupling function to explicitly condition vision prompts on their language counterparts, which acts as a bridge between the two modalities.

Nevertheless, there are two main challenges currently associated with efficient transfer learning for multimodal models. Firstly, existing multimodal transfer learning techniques~\citep{sung2022vl, lu2024uniadapter} are primarily limited to fine-tuning visual-language models, focusing only on the interaction between these two modalities and failing to extend to multimodal models with additional modalities. Secondly, the existing transfer learning strategies for multimodal models exhibit limited exploitation of interactions between modalities and lack efficiency when applied to models with multiple modalities. Specifically, existing multimodal efficient transfer learning strategies focus on how to fuse vector representations from different modalities rather than sequences of vector representations from different modalities, overlooking the interactions of time dimensions of various modalities. 

Based on the above observations, in this paper, we propose the lo\textbf{W}-r\textbf{a}nk seque\textbf{n}ce multimo\textbf{d}al adapt\textbf{er} (\textbf{Wander}) for efficient multimodal transfer learning. Wander enables fine-grained token-level interactions between sequences of different modalities in a parameter-efficient way. Specifically, motivated by the outer product fusion and low-rank decomposition, we fuse the information from different modalities in an element-wise and token-level way. Furthermore, we utilize CP decomposition to factorize tensors into low-rank components and achieve a substantial reduction in the number of parameters. We conduct extensive experiments on four datasets with different numbers of modalities. The performances demonstrate the effectiveness and efficiency of Wander.
In summary, our contributions are as follows:
\begin{itemize}
   \item We propose the low-rank sequence multimodal adapter that can be applied to situations with any number of modalities in a parameter-efficient way.
   \item Wander enables fine-grained token-level interactions between sequences of different modalities.
   \item Wander outperforms other efficient transfer learning methods consistently with fewer parameters on all the datasets.
\end{itemize}

\section{Related Work}
\textbf{Efficient Transfer Learning.} Efficient transfer learning offers a way to fine-tune the model with much fewer parameters and lower computational resources than full finetuning while achieving comparable performance. Efficient transfer learning can be mainly divided into two groups: additive fine-tuning methods and LoRA-based methods. Additive fine-tuning methods, which can be further divided into adapter methods~\citep{houlsby2019parameter, sung2022vl,lu2024uniadapter} and prompt methods~\citep{liu2023gpt, li2021prefix, vu2021spot, Wang2023ParameterefficientTO,guo2024multimodal, yan2024low}, add extra modules or parameters to the model and only train these added parameters while freezing the large pre-trained model. LoRA-based methods~\citep{hu2021lora, dettmers2024qlora, li2024loftq, chen2024longlora}, which are also referred to as reparameterization methods, construct low-rank weight matrices, add them to existing weights and only train these low-rank matrices. For multimodal models, adapter-based methods and prompt learning are popular. \citet{khattak2023maple} propose a coupling function to explicitly condition vision prompts on their language counterparts. \citet{lu2024uniadapter} propose a knowledge-sharing adapter design which enables efficient adaptation with cross-modal representations. \citet{Wang2023ParameterefficientTO} propose a prompt framework which utilizes mode approximation to implement multimodal efficient transfer learning. However, these methods are designed for vision-language models and fail to extend to situations where there are more than two modalities. Besides, these methods focus on sequence coarse-grained features, exhibiting limited exploitation of interactions between modalities. In contrast, our method can be applied to tasks with any number of modalities with token-level and fine-grained interactions between modalities.

\noindent\textbf{Multimodal Fusion.} Multimodal fusion can be divided into early fusion, late fusion and intermediate fusion. Early fusion methods~\citep{liu2023bevfusion, liang2022bevfusion} integrate multimodal features through concatenation or a simple function such as averaging before inputting into the models. Late fusion methods~\citep{tsai2019multimodal} use data from different modalities independently followed by fusion at a decision-making stage. Intermediate fusion methods~\citep{zadeh2017tensor, liu2018efficient, perez2019mfas, joze2020mmtm} allow data fusion at different stages of model training. For example, \citet{zadeh2017tensor} use the outer product to model element-wise features between modalities at different stages of the model. Compared with intermediate fusion strategies, early fusion and late fusion methods can not model inter-modality information effectively and are not as flexible as intermediate methods. However, intermediate methods have more parameters and existing intermediate methods focus on sequence-level features. In contrast, our method can model token-level features between sequences from different modalities in a parameter-efficient way by using low-rank factors.

\section{Methodology}\label{s3}
In this section, we will introduce our lo\textbf{W}-r\textbf{a}nk seque\textbf{n}ce multimo\textbf{d}al adapt\textbf{er} (\textbf{Wander}) for efficient multimodal transfer learning. We will first introduce preliminaries in Section~\ref{s30}. Then, we will make a simple analysis of the outer product fusion in Section~\ref{s31} before introducing our Wander architecture in Section~\ref{s32}.

\subsection{Preliminaries}\label{s30}
\noindent\textbf{Adapter Tuning.} Tuning with adapter~\citep{houlsby2019parameter} modules involves adding a small number of new parameters to a model. An adapter module consists of a feedforward down-projection layer, a nonlinearity and an up-projection layer. Besides, a skip connection is added. We can represent the adapter module as:
\begin{equation}
    \texttt{Adapter}(x) = x + \texttt{Up}(\texttt{Nonlinear}(\texttt{Down}(x)))
\end{equation}
where $\texttt{Up}(\cdot), \texttt{Nonlinear}(\cdot)$ and $\texttt{Down}(\cdot)$ represent the up-projection layer, nonlinear function and down-projection layer, respectively.

\noindent\textbf{CP decomposition.}  The CANDECOMP/PARAFAC or canonical polyadic (CP) decomposition factorizes a tensor into a sum of outer products of vectors. Given an $N$-dimensional tensor $\mathcal X\in\mathbb R^{d_1\times d_2\times\cdots\times d_N}$, it can be represented as a combination of tensors:
\begin{equation}
    \mathcal X = \sum_{r=1}^R \bigotimes_{n=1}^N a_n^r
\end{equation}
where $R$ is the rank and $a_n^r\in\mathbb R^{d_n}$. $\bigotimes_{n=1}^N$ is the tensor outer product operation over a set of vectors indexed by $n$.

\subsection{Outer Product Multimodal Fusion}\label{s31}
Multimodal fusion allows interaction between different modalities. Early fusion and late fusion are two common strategies. However, these methods simply integrate the inputs or outputs from different modalities using weighted averaging or several MLP layers, which can not model inter-modality interactions effectively~\citep{liu2018efficient}. Therefore, more intermediate approaches~\citep{zadeh2017tensor, liu2018efficient, perez2019mfas, joze2020mmtm} are proposed. One notable category of methods computes the outer product between unimodal representations~\citep{zadeh2017tensor} which has shown great success. Given $M$ modalities, we denote them as $m_1, m_2, \cdots, m_M$ and the representation of each modality as $h_1, h_2, \cdots, h_M$. The unimodal representation $h_i\in \mathbb R^{d_i}$ where $i=1,2,\cdots, M$ and $d_i$ is the dimension. Then outer product method integrates these representations into a multimodal representation $H$ as follows:
\begin{equation}\label{e1}
    H = \bigotimes_{m=1}^Mh_m
\end{equation}
where $\bigotimes_{m=1}^M$ is the tensor outer product operation over a set of vectors indexed by $m$. After outer product operation, the multimodal representation $H\in\mathbb R^{d_1\times d_2\times \cdots\times d_M}$ is projected into a vector representation $\tilde{H} $ using a linear layer:
\begin{equation}\label{e2}
    \tilde{H} = \mathbf{W}\cdot H + b = \mathbf{W}\cdot \bigotimes_{m=1}^Mh_m + b, \quad\tilde{H}\in \mathbb R^{d_h}
\end{equation}
where $\mathbf W$ is the weight matrix and $b$ is the bias of the linear layer. $d_h$ represents the dimension of the projected multimodal vector. The weight matrix $\mathbf W\in \mathbb R^{d_1\times d_2\times \cdots\times d_M\times d_h}$. 

Compared with other fusion methods such as late fusion methods, outer product operation enables element-wise interactions between different modalities, thus achieving better results. However, as Equation~\ref{e1} and \ref{e2} shows, outer product operation needs to explicitly calculate the high-dimensional tensor $H$ and needs a high-dimensional matrix $\mathbf W$ to project the multimodal representation $H$ into the vector $\tilde{H}$, where the number of parameters will increase exponentially with the number of modalities which needs lots of computational resources.

\subsection{Low-rank Sequence Multimodal Adapter}\label{s32}
\subsubsection{Motivation}
\begin{figure}
    \begin{center}
    \subfigure[Vector Fusion]{\includegraphics[width=0.35\columnwidth]{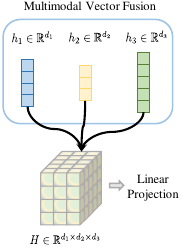}}\hspace{3pt}
    \subfigure[Sequence Fusion]{\includegraphics[width=0.62\columnwidth]{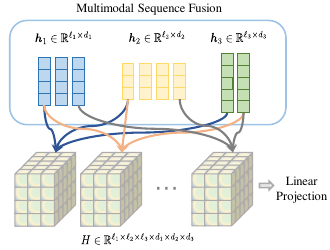}}
    \caption{The difference between vector fusion and sequence fusion in their original outer product form. Sequence fusion enables token-level interactions between modalities. We take three modalities as an example.}
    \label{diff}
    \end{center}
    \vskip -0.2in
\end{figure}

As aforementioned, existing efficient transfer learning methods for multimodal tasks have two main limitations. On the one hand, these methods~\citep{lu2024uniadapter, gao2024clip, zhang2021tip, sung2022vl} are designed for vision-language models and can not be extended to situations where there are more than two modalities. On the other, they focus on integrating vectors from different modalities rather than sequences of vectors which are fine-grained features (Figure~\ref{diff}). Motivated by outer product fusion introduced in Section~\ref{s31} and the low-rank multimodal fusion method~\citep{liu2018efficient}, we focus on addressing the two limitations and propose the Wander.

As discussed in Section~\ref{s31}, outer product fusion can model inter-modality interactions effectively because it enables element-wise calculation between modality vectors. However, it can not be applied to efficient transfer learning for multimodal tasks due to the need for heavy computational resources. Besides, it only models one vector representation $h\in \mathbb R^d$ where $d$ is the dimension of the representation, not sequences of vector representations from different modalities. Therefore, we need to make outer product fusion more efficient and enable it to deal with sequences of representations. To explain the low-rank sequence multimodal adapter more clearly, we first introduce low-rank single vector fusion before the sequence vector fusion.

\subsubsection{Vector fusion}\label{s321}
We first start by introducing the low-rank fusion of single vector representations from different modalities. According to Equation~\ref{e2}, we denote the linear projection matrix as $\mathbf W_h\in\mathbb R^{d_1\times d_2\times \cdots\times d_M\times d_h}$ which is an (M+1)-order tensor. We partition $\mathbf W_h$ into $\mathbf W_h^k\in\mathbb R^{d_1\times d_2\times \cdots\times d_M}, k=1,2,\cdots, d_h$. Then according to CP decomposition which factorizes a tensor into a sum of component rank-one tensors, we can transform $\mathbf W_h^k$ into a low-rank form:
\begin{equation}
    \mathbf W_h^k = \sum_{r=1}^R\bigotimes_{m=1}^M w_{h,m,k}^r, w_{h,m,k}^r\in\mathbb R^{d_m}
\end{equation}
where $R$ is a positive integer denoting the rank of the tensor. In other words, we can reconstruct $\mathbf W_k$ with these rank-one tensors $\{\{w_{h,m,k}^r\}_{m=1}^M\}_{r=1}^R$. We regroup these tensors into $M$ modality rank-one tensors. For a specific modality $m$, we can denote its decomposition factors as $\mathbf w_m=[\mathbf w_{h,m}^1, \mathbf w_{h,m}^2, \cdots, \mathbf w_{h,m}^R]$ where $\mathbf w_{h,m}^r=[w_{h,m,1}^r, w_{h,m,2}^r, \cdots, w_{h,m,d_h}^r]$ and $\mathbf w_{h,m}^r\in\mathbb R^{d_h\times d_m}$. Therefore, we can write the linear matrix $\mathbf W$ as follows:
\begin{equation}\label{e4}
    \mathbf W_h = \sum_{r=1}^R\bigotimes_{m=1}^M \mathbf w_{h,m}^r
\end{equation}
Then, based on the properties of outer products, summation, and element-wise multiplication, Equation~\ref{e2} can be rewritten as:
\begin{equation}\label{e5}
    \begin{aligned}
        \tilde{H} &= \sum_{r=1}^R\left (\bigotimes_{m=1}^M \mathbf w_{h,m}^r \cdot \bigotimes_{m=1}^M h_m\right ) + b_h\\
        &= \myhardamard_{m=1}^M \left[\sum_{r=1}^R\mathbf{w}_{h,m}^{r}\cdot h_m\right] +b_h
    \end{aligned}
\end{equation}
where $\Lambda_{m=1}^M$ denotes the element-wise multiplication over a sequence of vectors indexed by $m$. Compared with Equation~\ref{e2}, Equation~\ref{e5} does not compute the outer product explicitly which reduces the computational complexity. Besides, the parameter matrix $\mathbf W_h\in\mathbb R^{d_1\times d_2\times \cdots\times d_M\times d_h}$ is now transformed into $R$ 2-order matrix $\mathbf w_{h,m}^r\in\mathbb R^{d_h\times d_m}$, indicating a substantial reduction in the number of parameters.

\begin{figure}
    \begin{center}
    \centerline{\includegraphics[width=0.9\columnwidth]{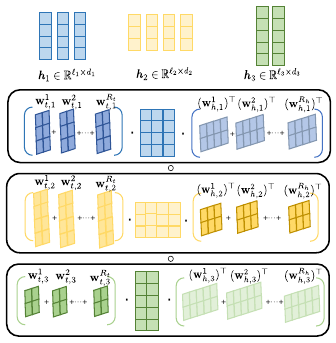}}
    \caption{The illustration of low-rank sequence fusion. We use three modalities as an example. $\circ$ denotes element-wise multiplication.}
    \label{sf}
    \end{center}
    \vskip -0.2in
\end{figure}

\subsubsection{Sequence fusion}\label{s322}
Vector fusion overlooks the fact that the Transformer outputs sequences of vectors which prevents it from modeling token-level and fine-grained features. Therefore, we propose sequence fusion to model these features in Transformers. The difference between vector fusion and sequence fusion is shown in Figure~\ref{diff}. Given multimodal representations $\boldsymbol h_1, \boldsymbol h_2, \cdots, \boldsymbol h_M$ where $M$ denotes the number of modalities and each representation $\boldsymbol h_m\in\mathbb R^{\ell_m\times d_m}, m=1, 2,\cdots, M$ consists of a sequence of vectors outputted by the Transformer layer. $d_m$ is the dimension of the vector and $\ell_m$ is the sequence length. Concretely, $\boldsymbol h_m = [h_m^1, h_m^2, \cdots, h_m^{\ell_m}]$ where $h_m^i\in\mathbb R^{d_m}, i=1,2,\cdots,\ell_m$. Given a vector representation $h_m^i$, we want it to interact with all the representations from all the other modalities. Specifically, we use the following equation to represent the process:
\begin{equation}\label{e6}
    H_{i_1,i_2, \cdots, i_M} = \texttt{VF}(h_1^{i_1}, h_2^{i_2}, \cdots, h_M^{i_M}),\quad H_{i_1,i_2, \cdots, i_M}\in\mathbb R^{d_h}
\end{equation}
where $\texttt{VF}$ denotes the vector fusion process in Section~\ref{s321}, $d_h$ is the dimension and $h_m^{i_m}$ denotes the $i_m$-th representation in $\boldsymbol h_m, m=1,2,\cdots, M, i_m=1,2,\cdots, \ell_m$. $H_{i_1,i_2, \cdots, i_M}$ denotes the integrated representation of the corresponding representations from different modalities. From Equation~\ref{e6}, we can observe that we will do vector fusion $\prod_{m=1}^M \ell_m$ times to get the final representation $H_t\in\mathbb R^{\ell_1\times\ell_2\times\cdots\times\ell_M\times d_h}$. We can represent $H_t$ as follows:
\begin{equation}
     H_t = H\cdot \mathbf W_h = \bigotimes_{m=1}^M \boldsymbol h_m\cdot\mathbf W_h  
\end{equation}
where $\mathbf W_h\in\mathbb R^{d_1\times d_2\times\cdots\times d_M\times d_h}$. In the following context, we will omit the bias term for simplicity. Then following Equation~\ref{e2}, we use a linear layer to project $H_t$ into a sequence representation $\tilde H_t$:
\begin{equation}\label{e9}
    \tilde H_t = \mathbf W_t \cdot H_t = \mathbf W_t \cdot H \cdot \mathbf W_h, \quad\tilde H_t \in\mathbb R^{d_t\times d_h}
\end{equation}
where the matrix $\mathbf W_t\in\mathbb R^{d_t\times\ell_1\times\ell_2\times\cdots\times\ell_M}$ and $d_t$ is the integrated sequence length. However, this poses the same problem as Equation~\ref{e2}, where the number of parameters of $\mathbf W_t$ is large and we need to explicitly calculate the high dimensional tensor $H$. To reduce the number of parameters, we also use CP decomposition to transform the matrix $\mathbf W_t$ into a series of rank-one tensors. Firstly, we partition $\mathbf W_t$ into $\mathbf W_t^k\in\mathbb R^{\ell_1\times \ell_2\times\cdots\times\ell_M}, k=1,2,\cdots,d_t$. According to CP decomposition, we have:
\begin{equation}
    \mathbf W_t^k = \sum_{r=1}^R\bigotimes_{m=1}^M w_{t,m,k}^r, w_{t,m,k}^r\in\mathbb R^{\ell_m}
\end{equation}
Similarly, we regroup these tensors $\mathbf W_t^k$ into $M$ modality rank-one tensors. We denote modality $m$ decomposition factors as $\mathbf w_{t,m}=[\mathbf w_{t,m}^1, \mathbf w_{t,m}^2, \cdots, \mathbf w_{t,m}^R]$ where $\mathbf w_{t,m}^r=[w_{t,m,1}^r, w_{t,m,2}^r, \cdots, w_{t,m,d_t}^r]$ and $\mathbf w_{t,m}^r\in\mathbb R^{d_t\times \ell_m}$. Then, $\mathbf W_t$ can be rewritten as:
\begin{equation}\label{e11}
    \mathbf W_t = \sum_{r=1}^R\bigotimes_{m=1}^M \mathbf w_{t,m}^r
\end{equation}

Then, using Equation~\ref{e4} and \ref{e11}, we can rewrite Equation~\ref{e9} as follows:
\begin{equation}\label{e12}
    \begin{aligned}
        \tilde H_t &= \sum_{r_t=1}^{R_t}\bigotimes_{m=1}^M \mathbf w_{t,m}^{r_t}\left [\sum_{r_h=1}^{R_h} \left (\bigotimes_{m=1}^M \mathbf w_{h,m}^{r_h}\cdot \bigotimes_{m=1}^M \mathbf h_m^\top\right )\right ]^\top\\
        &=\sum_{r_t=1}^{R_t}\sum_{r_h=1}^{R_h}\left (\bigotimes_{m=1}^M \mathbf w_{t,m}^{r_t}\cdot\bigotimes_{m=1}^M \mathbf h_m\cdot\bigotimes_{m=1}^M (\mathbf w_{h,m}^{r_h})^\top \right )\\
        &=\myhardamard_{m=1}^M\left [\sum_{r_t=1}^{R_t}\sum_{r_h=1}^{R_h}\mathbf w_{t,m}^{r_t}\cdot\mathbf h_m\cdot(\mathbf w_{h,m}^{r_h})^\top\right ]
    \end{aligned}
\end{equation}
where $\mathbf w_{h,m}^{r_h}\in\mathbb R^{d_h\times d_m}, \mathbf h_m\in\mathbb R^{\ell_m\times d_m}$ and $\mathbf w_{t,m}^{r_t}\in\mathbb R^{d_t\times \ell_m}$, $\mathbf w^\top$ represents the transpose of $\mathbf w$, and $R_t, R_h$ denotes the rank in Equation~\ref{e4} and \ref{e11}, respectively. From Equation~\ref{e12}, we can observe that it is no longer necessary to calculate $\bigotimes_{m=1}^M \boldsymbol h_m$ explicitly and the original high-dimensional matrices $\mathbf W_h$ and $\mathbf W_t$ are transformed into several low-rank 2D matrices. Figure~\ref{sf} presents the illustration of low-rank sequence fusion.

\begin{figure}
    \begin{center}
    \centerline{\includegraphics[width=0.88\linewidth]{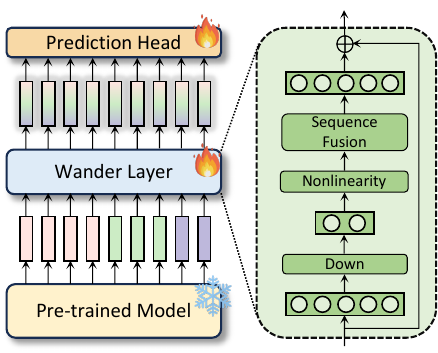}}
    \caption{The overall architecture of Wander and its integration with the pre-trained model. \textbf{Left:} We add Wander to the pre-trained model for fine-tuning. \textbf{Right:} Wander consists of a linear down-projection layer (Down),  a nonlinear function, sequence fusion and a skip connection.}
    \label{fover}
    \end{center}
    \vskip -0.2in
\end{figure}

\subsubsection{Wander Architecture}
Through Equation~\ref{e12}, we can integrate features from different modalities in a sequence token-level and parameter-efficient way, which can be extended to any number of modalities scenarios. Figure~\ref{fover} presents the overall architecture of Wander. Sequence Fusion is the main component of Wander, which enables sequence token-level interactions between modalities in a parameter-efficient way. Specifically, we denote the output of the pre-trained Transformer as $\boldsymbol h_1, \boldsymbol h_2, \cdots, \boldsymbol h_M$ where $M$ denotes the number of modalities and $\boldsymbol h_m\in\mathbb R^{\ell_m\times d_m}, m=1,2,\cdots, M$. For simplicity, we denote $\boldsymbol h = [\boldsymbol h_1, \boldsymbol h_2, \cdots, \boldsymbol h_M]$. Then, we can represent the process of Wander as:
\begin{equation}
    \texttt{Wander}(\boldsymbol h) = \boldsymbol h + \texttt{SF}(\texttt{Nonlinear}(\texttt{Down}(\boldsymbol h)))
\end{equation}
where $\texttt{Down}(\cdot)$ is the linear layer, $\texttt{SF}(\cdot)$ is the sequence fusion and $\texttt{Nonlinear}(\cdot)$ is the nonlinear function such as ReLU. Different from the original adapter module~\citep{houlsby2019parameter}, Wander does not have an up-projection layer because the sequence fusion has a projection layer internally. Therefore, we discard the original explicit up-projection layer, which makes Wander more parameter-efficient. Besides, the down-projection layer and sequence fusion are both linear transformations and thus we add a nonlinear function. In this paper, we use the same nonlinear function as the pre-trained model. In the fine-tuning stage, we freeze the pre-trained model and only train the Wander module and the prediction head which is usually a linear layer.

\section{Experiments}
\subsection{Experimental Settings}
\noindent\textbf{Datasets and Evaluation Metrics.} We evaluate Wander on four different downstream tasks with different numbers of modalities, including UPMC-Food 101~\citep{7169757} (2 modalities), CMU-MOSI~\citep{Zadeh2016MultimodalSI} (3 modalities), IEMOCAP~\citep{Busso2008IEMOCAPIE} (3 modalities), MSRVTT~\citep{7780940} (7 modalities). Particularly, to evaluate Wander in situations with more modalities, we use extracted features which consist of seven modalities~\citep{gabeur2020mmt} for MSRVTT. 

\noindent\textbf{UPMC-Food 101} is a food classification dataset, which contains about 100,000 recipes for a total of 101 food categories. Each item in the dataset is represented by one image plus textual information. 

\noindent\textbf{CMU-MOSI} is a popular dataset for multimodal (audio, text and video) sentiment analysis. 
These videos are carefully selected from YouTube and divided into 2,199 segments. Each segment is manually annotated with a sentiment score ranging from strongly negative to strongly positive (-3 to +3).

\noindent\textbf{IEMOCAP} is a multimodal (audio, text and video) emotion recognition dataset, which contains recorded videos from ten actors in five dyadic conversation sessions. Actors engaged in five different scenarios, performing selected emotional scripts and eliciting specific types of emotions (happiness, anger, sadness, frustration and neutral state). 
\begin{table}
    \centering
    \caption{Comparison of performance on the UPMC-Food 101 dataset. \#Tunable denotes the number of trainable parameters, including the added adapter and the task prediction head. ``d=" denotes the down projection dimension of the adapter.}
    \label{rfood}
    \resizebox{0.9\columnwidth}{!}{%
       \begin{tabular}{@{}lccc@{}}
        \toprule
        Method & \#Tunable & ACC & F1 \\ \midrule
        Full fine-tuning & 220M & \textbf{91.8} & \textbf{91.9} \\\midrule
        LoRA (r=64)~\citep{hu2021lora} & 26.4M & 88.7 & 88.7 \\
        Adapter (d=64)~\citep{houlsby2019parameter} & 24.9M & 88.4 & 84.3 \\
        Adapter (d=128)~\citep{houlsby2019parameter} & 50.1M & 89.2 & 89.3 \\
        P-tuning\citep{liu2023gpt} & 6.2M &83.1 & 83.0\\
        MaPLe~\citep{khattak2023maple} & 6.5M & 84.6 & 84.7 \\
        PMF~\citep{li2023efficient}&5.2M&89.1&89.1\\
        UniAdapter (d=128)~\citep{lu2024uniadapter} & 35.9M & 90.8 & 90.8 \\
        Aurora (d=128)~\citep{Wang2023ParameterefficientTO}&\textbf{2.6M}&90.2&90.1\\\midrule
        Wander (d=64) & 3.1M & 91.1 & 91.1 \\
        Wander (d=128) & 4.9M & \textbf{91.8} & 91.8 \\ \bottomrule
        \end{tabular}%
    }
\end{table}

\begin{table}[h!]
    \centering
    \caption{Comparison of performance on IEMOCAP.}
    \label{riemo}
    \resizebox{0.9\columnwidth}{!}{%
        \begin{tabular}{@{}lccc@{}}
            \toprule
            Method & \#Tunable & ACC & F1 \\ \midrule
            Full fine-tuning & 80M & \textbf{74.8} & 74.3 \\\midrule
            LoRA (r=16)~\citep{hu2021lora} & 4.0M & 71.5 & 71.1 \\
            Adapter (d=16)~\citep{houlsby2019parameter} & 4.0M & 70.8 & 71.7 \\
            Adapter (d=64)~\citep{houlsby2019parameter} & 15.8M & 72.0 & 71.7 \\
            P-tuning\citep{liu2023gpt} & 5.0M & 72.5 & 72.4 \\
            MaPLe~\citep{khattak2023maple} & 5.2M & 73.2 & 73.1 \\
            PMF~\citep{li2023efficient}&4.8M&72.9&72.6\\\midrule
            Wander (d=16) & \textbf{0.3M} & 74.2 & 73.8 \\
            Wander (d=64) & 1.0M & 74.7 & \textbf{74.4} \\ \bottomrule
        \end{tabular}%
    }
    \vskip -0.1in
\end{table}

\begin{table*}
    \centering
    \caption{Comparison of performance on the CMU-MOSI dataset.}
    \label{rmosi}
    \resizebox{0.62\linewidth}{!}{%
    \begin{tabular}{@{}lcccccc@{}}
    \toprule
    Method & \#Tunable & ACC-2 & F1 & ACC-7 & MAE & Corr \\ \midrule
    Full fine-tuning & 80M & 82.6 & 82.5 & 33.2 & 0.93 & 0.70 \\\midrule
    LoRA (r=16)~\citep{hu2021lora} & 4.0M & 81.2 & 81.0 & 29.3 & 0.96 & 0.67 \\
    Adapter (d=16)~\citep{houlsby2019parameter} & 3.9M & 81.1 & 80.9 & 29.4 & 0.97 & 0.66 \\
    Adapter (d=64)~\citep{houlsby2019parameter} & 15.7M & 81.6 & 81.6 & 30.1 & 0.95 & 0.68 \\
    P-tuning~\citep{liu2023gpt} & 5.0M & 81.3 & 81.2 & 28.4 & 0.98 & 0.66 \\
    MaPLe~\citep{khattak2023maple} & 5.2M & 82.1 & 82.0 & 31.2 & 0.95 & 0.68 \\
    PMF~\citep{li2023efficient}&4.8M&81.9&81.8&30.9&0.95&0.67\\\midrule
    Wander (d=16) & \textbf{0.3M} & 82.5 & 82.4 & 32.8 & 0.93 & 0.68 \\
    Wander (d=64) & 0.9M & \textbf{83.2} & \textbf{82.9} & \textbf{33.6} & \textbf{0.92} & \textbf{0.71} \\ \bottomrule
    \end{tabular}%
    }
\end{table*}

\begin{table*}[]
    \centering
    \caption{Comparison of performance on the MSRVTT dataset.}
    \label{rmsrvtt}
    \resizebox{0.76\linewidth}{!}{%
    \begin{tabular}{@{}lccccc|cccc@{}}
    \toprule
    \multirow{2}{*}{Method} & \multirow{2}{*}{\#Tunable} & \multicolumn{4}{c}{Text$\rightarrow$Video} & \multicolumn{4}{c}{Video$\rightarrow$Text} \\ \cmidrule(l){3-6} \cmidrule(l){7-10} 
     &  & R@5 & R@10 & MdR & MnR & R@5 & R@10 & MdR & MnR \\ \midrule
    Full fine-tuning & 134M & \textbf{57.2} & 69.3 & 4.0 & \textbf{22.4} & 57.8 & 68.5 & 4.0 & \textbf{20.1} \\\midrule
    LoRA (r=16)~\citep{hu2021lora} & 5.8M & 53.8 & 66.9 & 5.0 & 27.3 & 54.1 & 66.2 & 5.0 & 24.4 \\
    Adapter (d=8)~\citep{houlsby2019parameter} & 4.6M & 51.2 & 64.2 & 5.0 & 29.6 & 51.6 & 63.9 & 5.0 & 26.0 \\
    Adapter (d=16)~\citep{houlsby2019parameter} & 9.2M & 53.1 & 66.4 & 5.0 & 27.4 & 53.8 & 65.8 & 5.0 & 24.8 \\
    P-tuning~\citep{liu2023gpt} & 1.0M & 54.1 & 67.3 & 4.0 & 26.9 & 54.8 & 66.9 & 4.0 & 23.9 \\
    MaPLe~\citep{khattak2023maple} & 1.1M & 55.3 & 68.2 & 4.0 & 25.8 & 56.1 & 67.8 & 4.0 & 23.6 \\
    PMF~\citep{li2023efficient}&1.0M&54.6&67.8&4.0&26.4&55.3&67.2&4.0&23.7\\\midrule
    Wander (d=8) & \textbf{0.4M} & 56.4 & \textbf{69.4} & 4.0 & 22.8 & 57.3 & \textbf{68.9} & 4.0 & 21.4 \\
    Wander (d=16) & 0.8M & \textbf{57.2} & \textbf{69.4} & 4.0 & \textbf{22.4} & \textbf{58.0} & \textbf{68.9} & 4.0 & \textbf{20.1} \\ \bottomrule
    \end{tabular}%
    }
    \vskip -0.1in
\end{table*}

\noindent\textbf{MSRVTT} is characterized by unique properties including large-scale clip-sentence pairs, comprehensive video categories, diverse video content and descriptions, as well as multimodal audio and video streams. MSRVTT consists of 10,000 video clips from 20 categories, and each video clip is annotated with 20 English sentences. To validate Wander in situations with more modalities, we use extracted features which have seven modalities. Specifically, following previous work~\citep{gabeur2020mmt}, we extract motion, audio, scene, ocr, face, speech and appearance features. 

For UPMC-Food 101 and IEMOCAP, we use binary accuracy (ACC) and F1 score (F1) to evaluate the performance. For CMU-MOSI, we use binary accuracy (ACC-2), F1 score (F1), 7-class accuracy (ACC-7), mean absolute error (MAE, lower is better) and Pearson correlation (Corr, higher is better) to evaluate the performance. For MSRVTT, we use recall at rank N (R@N, higher is better), median rank (MdR, lower is better) and mean rank (MnR, lower is better) to evaluate the performance.

\noindent\textbf{Implementation Details.} For UPMC-Food 101, we use the pre-trained bert-base model~\citep{devlin2018bert} and pre-trained ViT~\citep{dosovitskiy2020image} as the backbone. For CMU-MOSI and IEMOCAP, we use stacked Transformer layers as the backbone. For MSRVTT, we use the video retrieval model MMT~\citep{gabeur2020mmt} as the backbone. For CMU-MOSI, IEMOCAP and MMT, we pre-train the model on the HowTo100M dataset~\citep{miech2019howto100m}. In the training process, we add Wander and the task prediction head to the backbone, keep the pre-trained backbone frozen and only train the Wander and the task prediction head. For all the datasets, we set $R_h$ and $R_t$ to 8 by default. For UPMC-Food 101, we set the batch size to 128 and use the AdamW optimizer with a StepLR scheduler where the initial learning rate is 2e-3, step is 30 and the decay rate $\gamma=0.1$. For CMU-MOSI and IEMOCAP, we set the batch size to 24 and use the Adam optimizer with a StepLR scheduler where the initial learning rate is 1e-3, step is 10 and the decay rate $\gamma=0.1$. For MSRVTT, we set the batch size to 64 and use the Adam optimizer with a StepLR scheduler where the initial learning rate is 3e-3, step is 1 and the decay rate $\gamma=0.96$.

\begin{table}
    \centering
    \caption{Comparison of Wander and its two original forms on CMU-MOSI. $-$c denotes we discard c in Wander.}
    \label{te}
    \resizebox{\linewidth}{!}{%
    \begin{tabular}{@{}lcccccc@{}}
    \toprule
    Method & \#Tunable & ACC-2 & F1 & ACC-7 & MAE & Corr \\ \midrule
    SF-OP & 3.5M & 81.6 & 81.2 & 29.9 & 0.96 & 0.67 \\
    SF-VF & 1.2M & 81.5 & 81.3 & 29.6 & 0.96 & 0.67 \\\midrule
    Wander (d=16) & \textbf{0.1M} & 81.8 & 81.4 & 30.6 & 0.95 & 0.67 \\
    $\quad-$nonlinearity & 0.2M & 81.6 & 81.3 & 29.8 & 0.96 & 0.67 \\
    $\quad-$residual & 0.1M & 81.5 & 81.2 & 29.6 & 0.96 & 0.67 \\\midrule
    Wander (d=32) & 0.1M & \textbf{82.2} & \textbf{82.1} & \textbf{31.0} & \textbf{0.94} & \textbf{0.68} \\ 
    $\quad-$nonlinearity & 0.2M & 82.1 & 82.1 & 31.1 & 0.94 & 0.68 \\
    $\quad-$residual & 0.2M & 81.9 & 81.7 & 30.9 & 0.94 & 0.68 \\\bottomrule
    \end{tabular}%
    }
    \vskip -0.1in
\end{table}

\subsection{Comparison with State-of-the-arts}
For UPMC-Food 101 which has only two modalities, we compare Wander with state-of-the-art efficient transfer learning methods, including unimodal methods LoRA~\citep{hu2021lora}, Adapter~\citep{houlsby2019parameter}, P-tuning~\citep{liu2023gpt} and vision-language transfer learning methods MaPLe~\citep{khattak2023maple}, UniAdapter~\citep{lu2024uniadapter}, PMF~\citep{li2023efficient} and Aurora~\citep{Wang2023ParameterefficientTO}. Particularly, UniAdapter is based on BLIP~\citep{li2022blip}. Therefore, we modify it slightly for our backbones. For datasets with more modalities, we compare Wander with LoRA~\citep{hu2021lora}, Adapter~\citep{houlsby2019parameter}, P-tuning~\citep{liu2023gpt}, MaPLe~\citep{khattak2023maple} and PMF~\citep{li2023efficient}. We modify the prompts design of MaPLe and PMF to fit more modality situations. For MaPLe, we use a cascade structure to condition the prompts between modalities.

Table~\ref{rfood}, \ref{riemo}, \ref{rmosi} and \ref{rmsrvtt} presents the results on the four datasets with different numbers of modalities. We can observe that Wander consistently outperforms other efficient transfer learning methods and can achieve competitive results as full fine-tuning models. Besides, Wander has fewer parameters than other methods. Particularly, on the CMU-MOSI and MSRVTT datasets which have three and seven modalities, Wander outperforms the full fine-tuning models, indicating its superiority on situations where there are more than two modalities. When trained with more tunable parameters (\textit{i.e.} the larger down projection dimension), Wander can further boost the performance. The results in the tables demonstrate that Wander can enable sufficient interactions between modalities in a token-level and parameter-efficient way.

\subsection{Effectiveness of Wander}
To evaluate the efficiency and effectiveness of Wander, we compare Wander with its original forms. Specifically, we denote the outer product form of sequence fusion in Equation~\ref{e9} as SF-OP and the vector fusion form of sequence fusion in Equation~\ref{e6} as SF-VF. The results on the CMU-MOSI dataset of Wander and its two original forms are presented in Table~\ref{te}. We reduce the dimension of the features because the weight matrices $\mathbf W_h\in \mathbb R^{d_1\times d_2\times \cdots\times d_M\times d_h}$ and $\mathbf W_t\in\mathbb R^{d_t\times\ell_1\times\ell_2\times\cdots\times\ell_M}$ in SF-OP are very high-dimensional matrices. From the table, we can observe that SF (row 4) can rival its original forms SF-OP and SF-VF. This indicates the effectiveness of the low-rank decomposition of the high-dimensional matrices $\mathbf W_h$ and $\mathbf W_t$ without performance degradation. Moreover, Wander (d=16) and Wander (d=32) both outperform SF-OP and SF-VF, indicating the effectiveness of the design of Wander. SF-OP and SF-VF are both linear operations whereas in Wander we add nonlinearity and residual block to further enhance the performance.

\begin{table}[]
    \centering
    \caption{Computational analysis of low-rank sequence fusion and its two original forms on CMU-MOSI.}
    \label{comana}
    \resizebox{0.74\columnwidth}{!}{%
    \begin{tabular}{@{}lccc@{}}
    \toprule
    Method & GPU time (s) & Memory (MB) & FLOPs \\ \midrule
    SF-OP & 9.16 & 3300 & 0.003 \\
    SF-VF & 7.87 & 622 & 0.002 \\
    SF & \bf{0.015} & \bf{348} & \bf{0.002} \\ \bottomrule
    \end{tabular}%
    }
\end{table}

\begin{table}[]
    \centering
    \caption{Benefits of sequence fusion on CMU-MOSI. VF denotes the vector fusion and SF denotes the sequence fusion. The down-projection dimensions are both 64.}
    \label{ab1}
    \resizebox{0.85\columnwidth}{!}{%
    \begin{tabular}{@{}cccccc@{}}
    \toprule
    Method & ACC-2 & F1 & ACC-7 & MAE & Corr \\ \midrule
    Wander (VF) & 80.9 & 80.4 & 29.3 & 0.98 & 0.67 \\
    Wander (SF) & \textbf{83.2} & \textbf{82.9} & \textbf{33.6} & \textbf{0.92} & \textbf{0.71} \\ \bottomrule
    \end{tabular}%
    }
    \vskip -0.1in
\end{table}

\begin{figure}[h!]
    \centering
    \centerline{\includegraphics[width=0.95\linewidth]{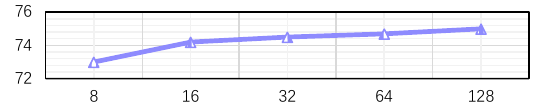}}
    \caption{The impact of $d$ on the performance on IEMOCAP.}
    \label{dim}
    \vskip -0.2in
\end{figure}

\subsection{Cost Analysis}
For the efficiency of Wander, from Table~\ref{te}, because we reduce the dimension of the features to enable the training of SF-OP and SF-VF, we fail to observe a substantial reduction in the number of parameters. However, mathematically, according to Equation~\ref{e9}, it is easy to observe that the complexity of the number of parameters of SF-OP is $\mathcal O(d_h\times\prod_{i=1}^{m}d_m+d_t\times\prod_{i=1}^{m}\ell_m)$. Similarly, the complexity of SF-VF is $\mathcal O(R_h\times d_h\times\sum_{i=1}^{m}d_m+d_t\times\prod_{i=1}^{m}\ell_m)$ and the complexity of Wander is reduced to $\mathcal O(R_h\times d_h\times\sum_{i=1}^{m}d_m+R_t\times d_t\times\sum_{i=1}^{m}\ell_m)$. For example, if there are three modalities and the shape of each modality is $(10, 768)$ (\textit{i.e.} the length is 10 and the dimension is 768), then the number of parameters of SF-OP is around 348B while that of Wander is only around 14M. This fully demonstrates the efficiency of Wander. Additionally, we present the comparison of GPU time, memory and FLOPs in Table~\ref{comana}. We use a batch size of 24 for all methods and still reduce the dimension of features. Despite the reduction in dimension, we can still observe a substantial reduction in GPU time and memory, indicating the effectiveness of our low-rank decomposition.

\subsection{Ablation Study}
\noindent\textbf{Vector Fusion and Sequence Fusion.} To validate the core component of Wander, we compare the vector fusion and our proposed sequence fusion in Table~\ref{ab1} on the CMU-MOSI dataset. Specifically, we choose the first token (CLS token)~\citep{dosovitskiy2020image} of the Transformer as the fusion vector in vector fusion. From the table, we can observe that our proposed sequence fusion significantly outperforms the vector fusion, indicating the effectiveness and superiority of our sequence fusion method.

\begin{figure}
    \centering
    \subfigure[The impact of $R_h$]{\includegraphics[width=0.49\columnwidth]{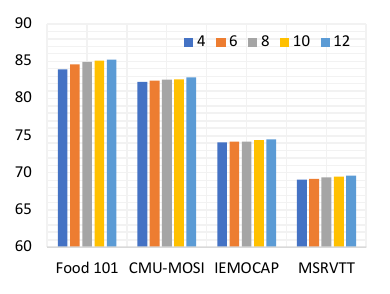}}\hspace{2pt}
    \subfigure[The impact of $R_t$]{\includegraphics[width=0.49\columnwidth]{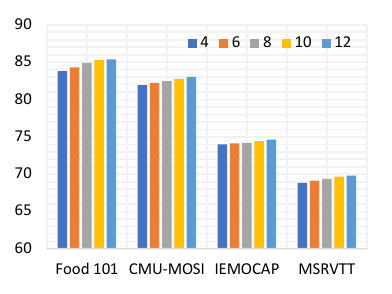}}
    \caption{The impact of the rank of CP decomposition on the performance. We report binary accuracy for UPMC-Food 101, CMU-MOSI and IEMOCAP and R@10 for MSRVTT.}
    \label{rank}
    \vskip -0.15in
\end{figure}

\noindent\textbf{Rank $R_h$ and $R_t$.} To explore the impact of the CP decomposition rank on the performance of the model, we select different ranks and present the results in Figure~\ref{rank}. From the figure, we can observe that with the increase of the rank, the performance of the model slightly improves. Besides, compared to $R_h$, the increase of $R_t$ brings more improvement. However, the value of the rank will not affect the performance significantly, indicating the robustness of Wander.

\begin{table}
    \centering
    \caption{The impact of the pre-training datasets on the performance of the model on CMU-MOSI and IEMOCAP datasets.}
    \label{pretrain}
    \resizebox{0.82\linewidth}{!}{%
    \begin{tabular}{@{}c|c|cc|cc@{}}
    \toprule
    \multirow{2}{*}{Pre-training} & \multirow{2}{*}{Method} & \multicolumn{2}{c}{CMU-MOSI} & \multicolumn{2}{c}{IEMOCAP} \\ \cmidrule(l){3-4} \cmidrule(l){5-6} 
     &  & ACC & F1 & ACC & F1 \\ \midrule
    \multirow{3}{*}{HowTo100M} & Full fine-tuning & 82.6 & 82.5 & 74.8 & 74.3 \\
     & Wander(d=16) & 82.5 & 82.4 &  74.2 & 73.8 \\
     & Wander(d=64) & 83.2 & 82.9 & 74.7 & 74.4 \\\midrule\midrule
    \multirow{3}{*}{CMU-MOSEI} & Full fine-tuning & 83.3 & 83.2 & 75.3 & 74.9 \\
     & Wander(d=16) & 83.2 & 83.1 & 75.1 & 74.8 \\
     & Wander(d=64) & 83.6 & 83.4 & 75.6 & 75.3 \\ \bottomrule
    \end{tabular}%
    }
\end{table}

\noindent\textbf{Rank $d$.} In Figure~\ref{dim}, we explore the impact of $d$ on the performance of the model. As $d$ increases, the performance improvement decreases slowly. Empirically, we can set $d$ to around 1/8 of the original dimension. This will achieve a balance between performance and number of parameters.

\noindent\textbf{Pre-training Datasets.} To explore the impact of pre-training datasets on the performance of the model, we use two different datasets for pre-training and present the results on CMU-MOSI and IEMOCAP in Table~\ref{pretrain}. Specifically, CMU-MOSEI is a large multimodal sentiment analysis dataset. Therefore, we choose it as one of the pre-training datasets. From the table, we can observe that Wander can achieve good performance in both pre-training settings, indicating the universality of Wander.

\section{Conclusion}
In this paper, we address the two limitations of existing multimodal transfer learning methods: 1) existing methods focus on vision language transfer learning, failing to extend to situations with more modalities. 2) existing methods exhibit limited exploitation of interactions between modalities. Therefore, we propose the low-rank sequence multimodal adapter (Wander). Wander enables fine-grained token-level interactions between sequences of different modalities in a parameter-efficient way. We conduct extensive experiments on four datasets with different numbers of modalities. Wander outperforms state-of-the-art transfer learning methods consistently with fewer parameters, indicating its superiority.

\bibliography{aaai25}

\end{document}